\documentclass[conference]{IEEEtran}
\IEEEoverridecommandlockouts
\usepackage{cite}
\usepackage{amsmath,amssymb,amsfonts}
\usepackage{graphicx}
\usepackage{textcomp}
\usepackage{xcolor}
\usepackage{algorithm}
\usepackage{amsmath}
\usepackage[noend]{algpseudocode}
\usepackage{hyperref}
\usepackage{booktabs}
\usepackage{stfloats} 
\hypersetup{
    colorlinks=true,
    linkcolor=black,
    citecolor=black,
    urlcolor=blue,
    pdfborder={0 0 0}
}

\def\BibTeX{{\rm B\kern-.05em{\sc i\kern-.025em b}\kern-.08em
    T\kern-.1667em\lower.7ex\hbox{E}\kern-.125emX}}
\begin{document}

\title{ G-CSEA: A Graph-Based Conflict Set Extraction Algorithm for Identifying Infeasibility in Pseudo-Boolean Models\\
}

 \author{\IEEEauthorblockN{Kanishk Garg }
 \IEEEauthorblockA{ \textit{Sprinklr AI} \\
 Gurugram, India \\
 }
 \and
 \IEEEauthorblockN{Saranya D.}
 \IEEEauthorblockA{ \textit{Sprinklr AI} \\
 Gurugram, India }
 \and
 \IEEEauthorblockN{Sanal Kumar }
 \IEEEauthorblockA{\textit{Sprinklr AI}\\
 Gurugram, India \\
}
\and
 \IEEEauthorblockN{Saurabh Singh}
 \IEEEauthorblockA{ \textit{Sprinklr AI}\\
 Gurugram, India \\
}
\and
\IEEEauthorblockN{Anupam Purwar\IEEEauthorrefmark{1}}
\IEEEauthorblockA{\textit{Sprinklr AI} \\
Gurugram, India \\
}\thanks{\IEEEauthorrefmark{1}Corresponding Author: Anupam Purwar (e-mail: anupam.aiml@gmail.com, https://anupam-purwar.github.io/page/)}

}

\maketitle

\begin{abstract}
Workforce scheduling involves a variety of rule-based constraints—such as shift limits, staffing policies, working hour restrictions, and many similar scheduling rules—which can interact in conflicting ways, leading to infeasible models. Identifying the underlying causes of such infeasibility is critical for resolving scheduling issues and restoring feasibility. A common diagnostic approach is to compute \textit{Irreducible Infeasible Subsets} (IISs): minimal sets of constraints that are jointly infeasible but become feasible when any one is removed. We consider models formulated using pseudo-Boolean constraints with inequality relations over binary variables, which naturally encode scheduling logic. Existing IIS extraction methods such as Additive Deletion and QuickXplain rely on repeated feasibility checks, often incurring large numbers of solver calls. Dual ray analysis, while effective for LP-based models, may fail when the relaxed problem is feasible but the underlying pseudo-Boolean model is not. To address these limitations, we propose Graph-based Conflict Set Extraction Algorithm (G-CSEA) to extract a conflict set, an approach inspired by Conflict-Driven Clause Learning (CDCL) in SAT solvers. Our method constructs an implication graph during constraint propagation and, upon detecting a conflict, traces all contributing constraints across both decision branches. The resulting conflict set can optionally be minimized using QuickXplain to produce an IIS. We evaluate our approach on infeasible instances of workforce scheduling problems, comparing it with Additive Deletion, QuickXplain, Dual Ray with Deletion, and Elastic Filter with QuickXplain. Experimental results show that our conflict graph-based method produces compact conflict sets—typically reducing the constraint set by around 94\%—which, when passed to QuickXplain, substantially reduce the number of solver calls required for IIS extraction. Moreover, the method achieves lower IIS computation time than other techniques in most cases, while maintaining scalability across a range of model sizes. On average, the conflict graph + QuickXplain method achieves approximately 40\% reduction in IIS computation time compared to using QuickXplain alone. This makes G-CSEA a viable approach for diagnosing infeasibility in constraint-based scheduling systems.

\end{abstract}


\section{Introduction}
Workforce scheduling involves assigning employees to shifts over time while satisfying a wide range of operational constraints—such as limits on working hours, restrictions on consecutive workdays, day-offs, shift assignment rules, and other agent- or shift-specific requirements ~\cite{b10, b11}. These constraints are often modeled using pseudo-Boolean (PB) expressions: linear inequalities over Boolean variables, which directly encode binary scheduling decisions such as assigning an agent to a shift on a given day. When multiple scheduling policies interact—especially across different constraint types—the overall model may become infeasible due to conflicting requirements.

In such situations, simply reporting that the model is infeasible offers limited practical insight. Instead, it is crucial to identify \textit{Irreducible Infeasible Subsets} (IISs): minimal sets of constraints responsible for infeasibility, such that removing any single constraint restores satisfiability. IISs serve as a valuable tool for diagnosing infeasibility, debugging constraint models, and resolving policy-level conflicts in complex scheduling systems.

Early work on IIS computation primarily focused on deletion-based and slack-based techniques. Chinneck~\cite{b1} proposed practical heuristics such as deletion filtering, constraint reordering, constraint grouping, sensitivity analysis filters, and the elastic filter—which augments each constraint with a slack variable and minimizes the total slack to identify likely sources of infeasibility. While effective for continuous linear programs, these methods were not designed for, nor evaluated on, pseudo-Boolean constraint systems involving strictly Boolean variables.

Building on this, Guieu and Chinneck~\cite{b2} extended both the deletion filtering and additive techniques to mixed-integer linear programs (MILPs), incorporating hybrid additive-deletion strategies and dynamic constraint reordering. While their approach is broadly applicable to general MILP models, their experimental work focused primarily on mixed-integer cases and did not explore purely Boolean or pseudo-Boolean domains. This motivated us to adapt and evaluate deletion-style techniques for infeasible pseudo-Boolean scheduling problems.

To reduce the number of feasibility checks required during IIS computation, Junker~\cite{b3} proposed \textit{QuickXplain}, a general-purpose algorithm that employs a recursive divide-and-conquer strategy to isolate minimal conflict sets in constraint satisfaction problems. While QuickXplain has been widely adopted in domains such as model-based diagnosis and constraint explanation, it was not originally applied to infeasible pseudo-Boolean systems, and no empirical results were reported for such models.

Infeasibility analysis has also been studied through dual ray techniques, particularly the work of Gleeson and Ryan~\cite{b4}, which uses Farkas’ Lemma to identify minimal infeasible subsystems via unbounded dual rays. While theoretically elegant, this method applies only to continuous LPs and was not developed for pseudo-Boolean or binary-variable systems.

Conflict-driven techniques from SAT solving have also been extended to mixed-integer programming (MIP). Achterberg~\cite{b5} developed a conflict graph framework for the SCIP solver, which constructs implication graphs to derive conflict constraints—valid inequalities that help prune infeasible regions of the search space. The Galena solver, proposed by Chai and Kuehlmann~\cite{b7}, extends SAT-style conflict learning to pseudo-Boolean constraints by integrating DPLL-based search with cutting planes and watched literal propagation. Mexi et al.~\cite{b6} recently incorporated pseudo-Boolean reasoning into conflict analysis for MIP solvers through cutting-plane-based resolution methods, including division, saturation, and mixed integer rounding (MIR). Their approach enhances solver performance by generating stronger conflict constraints during search.

While these approaches significantly improve solver performance through more effective pruning and branching, they are designed primarily as internal search optimizations. None of them focus on identifying infeasibility in the original constraint model or extracting constraint-level explanations such as IISs.




In the SAT and CSP domains, Liffiton and Sakallah~\cite{b8} proposed the CAMUS framework for computing all Minimal Unsatisfiable Subsets (MUSes) of CNF Boolean formulas. Their method follows a two-phase strategy: first identifying Minimal Correction Subsets (MCSes), and then computing MUSes as minimal hitting sets of these MCSes. While general and influential, the approach is restricted to CNF clause sets and does not extend to pseudo-Boolean constraints involving cardinality or linear arithmetic.

Bailey and Stuckey~\cite{b9} proposed a general hitting set dualization approach for computing all Minimal Unsatisfiable Subsets (MUSes) in constraint satisfaction problems. While the method is complete, it can be computationally expensive even for extracting a single MUS, due to the cost of enumerating correction sets and solving the associated hitting set problem. 

To address these gaps, we introduce a conflict set extraction algorithm inspired by conflict-driven reasoning, tailored to pseudo-Boolean constraint systems. We also present experimental results on infeasible scheduling problems, comparing the performance of our approach against several established IIS extraction techniques.

\section{Methodology}
The following section contains details related to how the IIS extraction problem has been setup, followed by details on how the G-CSEA approach is inspired from conflict set extraction. More details on Conflict detection and analysis as well as MIR-based constraint learning are covered in sub-sections C and D.

\subsection{Problem Setup}
We consider the problem of identifying an \textit{Irreducible Infeasible Subset} (IIS) from an infeasible set of pseudo-Boolean constraints. Each constraint is of the form:

\[
\sum_{i \in I} a_i x_i \odot b, \quad x_i \in \{0,1\}, \quad a_i, b \in \mathbb{Z}, \quad \odot \in \{\leq, \geq, =\}
\]

where $x_i$ are Boolean decision variables, $a_i$ are integer coefficients, $b$ is the right-hand side constant, and $\odot$ denotes the constraint sense. These constraints naturally represent scheduling rules.
Given a finite set $\mathcal{C}$ of such constraints, we assume that $\mathcal{C}$ is infeasible—i.e., there is no assignment to the variables that satisfies all constraints simultaneously.

An \textbf{IIS} is a subset $\mathcal{C}' \subseteq \mathcal{C}$ such that: $\mathcal{C}'$ is infeasible, and every proper subset $\mathcal{C}'' \subsetneq \mathcal{C}'$ is feasible.

The goal is to compute such a minimal infeasible subset $\mathcal{C}'$, ideally with minimal computational overhead. Our focus is on developing a practical and scalable method for extracting such subsets efficiently from workforce scheduling models, where the number and complexity of constraints can be substantial.
\subsection{CDCL-Inspired Conflict Set Extraction}

Our approach follows a Conflict-Driven Clause Learning (CDCL)-style search procedure adapted to pseudo-Boolean constraints. The goal is to identify a subset of constraints responsible for infeasibility by tracking propagation and conflict dependencies.

The algorithm begins with an empty variable assignment and iteratively attempts to assign values to variables through propagation or decision-making. At each step, the algorithm performs unit propagation over all constraints. If a constraint becomes violated under the current partial assignment, a conflict is triggered, and conflict analysis is invoked.

If propagation leads to new variable assignments, all constraints are re-evaluated, and this process repeats until no further inferences can be made. If no conflict is detected and the assignment is still incomplete, the algorithm selects an unassigned variable and assigns it a value (0 or 1) as a decision. This increases the decision level and continues the propagation process.

When a conflict is detected, the implication structure is analyzed to identify all constraints that contributed to the conflict. These constraints are collected into a conflict set, and the algorithm backtracks to the latest decision level where unexplored assignments remain.

The process continues—propagating, deciding, analyzing conflicts, and backtracking—until either:
\begin{itemize}
  \item A satisfying assignment is found, indicating that the model is feasible, or
  \item All possible decision paths are exhausted and conflicts persist, in which case the algorithm returns the accumulated conflict set.
\end{itemize}

\vspace{1ex}
\noindent
\textit{High-Level Algorithm Outline:} The main control flow of the CDCL-based conflict set extraction algorithm is as follows:

\begin{enumerate}
    \item \textit{Propagation}:
    The algorithm performs unit propagation over all pseudo-Boolean constraints to extend the current partial assignment by checking for conflicts or inferring variable values using slack-based reasoning and implication rules. If any constraint is violated under the current assignment, a conflict is detected and conflict analysis is invoked. Whenever a variable is inferred, the constraint responsible is recorded for subsequent conflict analysis.

    \item \textit{Conflict Analysis} (see Section~\ref{sec:conflict_analysis}):
    When a conflict occurs, the implication graph is traversed to identify all constraints that contributed to the conflict. These are recorded as part of the conflict set. Additionally, the most recent decision variable involved in the conflict is identified for potential backtracking.

    \item \textit{Backtracking and Flipping:}  
    If both values (0 and 1) of the conflicting decision variable have been tried, the algorithm backtracks to the most recent decision level with unexplored branches. Otherwise, it flips the value of the current decision variable and resumes propagation.

    \item \textit{Decision Assignment:}  
    If no conflict is found and some variables remain unassigned, the algorithm selects the next unassigned variable and assigns it a decision value (0 or 1). The decision level is incremented, and the new assignment is pushed onto the decision stack.

    \item \textit{Termination:}  
    If a complete assignment is found that satisfies all constraints, the model is feasible and the algorithm returns \texttt{SAT}.  
    If all possible decision paths have been explored and conflicts persist, the model is declared \texttt{UNSAT}, and the collected conflict set is returned as the infeasibility explanation.
\end{enumerate}








\subsection{Conflict Detection and Analysis} \label{sec:conflict_analysis}

When a constraint is violated during propagation, a conflict is detected. The algorithm then performs conflict analysis to identify the subset of constraints responsible for the infeasibility under the current assignment.

To do this, we maintain an implication graph, where each implied variable is linked to the constraint that caused its assignment. Conflict analysis involves a backward traversal of this graph starting from the violated constraint:

\begin{itemize}
    \item For each variable in the conflict constraint, we trace its implication source—i.e., the constraint that forced its assignment.
    \item These source constraints are added to a traversal stack and explored recursively.
    \item If a variable was assigned via a decision (not implied), we record its decision level to identify where to backjump.
\end{itemize}

This traversal gathers all constraints that lie along implication paths contributing to the conflict, forming the conflict set. The most recent decision variable involved in the conflict is also identified and used to determine the appropriate backjump level.

After all paths are traversed, the collected constraints are stored as the conflict core—a non-minimal explanation of infeasibility. This set can optionally be minimized using QuickXplain to produce an Irreducible Infeasible Subset (IIS).

\subsection{Learning via MIR-based Constraint Derivation}

To strengthen the solver and reduce repeated exploration of the same conflicts, we incorporate an optional constraint learning mechanism based on the Mixed Integer Rounding (MIR) approach introduced by Mexi et al.~\cite{b6}. After a conflict is detected and the contributing constraints are identified, we apply an exact MIR-based resolution procedure to derive a new pseudo-Boolean constraint that generalizes the conflict.

This learned constraint captures the essence of the infeasibility in a compact, implying valid form. It can optionally be added back to the constraint set to improve future propagation and pruning. While not required for conflict set extraction or IIS computation, this learning step serves as a useful enhancement—particularly in large or structured infeasible systems where similar conflicts recur.

\begin{algorithm}
\footnotesize
\caption{CDCL-style Conflict Set Extraction}
\begin{algorithmic}[1]
\State \textbf{Input:} Pseudo-Boolean constraints
\State \textbf{Output:} Conflict core $\mathcal{C}$ or $\emptyset$ if feasible
\State $\mathcal{C} \gets \emptyset$, $tried\_decisions$ $\gets \emptyset$
\State Set decision level $\ell \gets 0$
\While{true}
  \State $conflict \gets$ \Call{Propagate}{}
  \If{$conflict$}
    \State $latest\_decision\_var \gets$ \Call{AnalyzeConflict}{$conflict$}
    \While{true}
      \If{$latest\_decision\_var = $ None}
        \State \Return $\mathcal{C}$ \Comment{Infeasible}
      \EndIf
      \State $tried \gets \text{$tried\_decisions$}[latest\_decision\_var] $

\State add latest assigned value of $latest\_decision\_var$ to $tried$
      \If{both 0 and 1 in $tried$}
        \State remove $latest\_decision\_var$ from $tried\_decisions$
        \If{previous decision exists}
          \State $latest\_decision\_var \gets$ previous decision
          \State \textbf{continue}
        \Else
          \State \Return $\mathcal{C}$ \Comment{All paths explored}
        \EndIf
      \Else
        \State \Call{Backtrack}{level of $latest\_decision\_var$}
        \State flip value of $latest\_decision\_var$
        \State $tried\_decisions$[$latest\_decision\_var$] $\gets$ $tried$
        \State \Call{Decide}{$latest\_decision\_var$, flipped value}
        \State \textbf{break}
      \EndIf
    \EndWhile
  \ElsIf{all variables assigned}
    \State \Return $\emptyset$ \Comment{Feasible}
  \Else
    \State choose next unassigned $x$
    \State \Call{Decide}{$x$, default value} \Comment{e.g., 1}
  \EndIf
\EndWhile
\end{algorithmic}
\end{algorithm}

\begin{algorithm}
\footnotesize
\caption{Analyze Conflict}
\begin{algorithmic}[1]
\State \textbf{Input:} Violated constraint $c$
\State \textbf{Output:} Latest decision variable involved in the conflict
\State Initialize stack with $c$
\State $visited\_constraints \gets \emptyset$
\State $latest\_decision\_var \gets$ None, $latest\_decision\_level \gets -1$
\While{stack not empty}
  \State $cname \gets$ stack.pop()
  \If{$cname \in visited\_constraints$}
    \State \textbf{continue}
  \Else
    \State add $cname$ to $visited\_constraints$
    \For{each variable $x$ in constraint $cname$}
      \If{$x$ was implied}
        \State $reason \gets$ constraint that implied $x$
        \State add $reason$ to stack
      \Else
        \State $y \gets$ decision level of $x$
        \If{$y > latest\_decision\_level$}
          \State $latest\_decision\_var \gets x$
          \State $latest\_decision\_level \gets y$
        \EndIf
      \EndIf
    \EndFor
  \EndIf
\EndWhile
\State Update global conflict core: $\mathcal{C} \gets \mathcal{C} \cup visited\_constraints$
\State $learned\_constraint \gets$ \Call{DeriveMIRConflict}{$c$}
\State add $learned\_constraint$ to the model
\State \Return $latest\_decision\_var$
\end{algorithmic}
\end{algorithm}

\newpage
\section{Results}

We evaluated the performance of our proposed Conflict Graph Analysis algorithm on a benchmark suite of 50 infeasible workforce scheduling instances. These models consist of binary decision variables and pseudo-Boolean constraints. For comparison, we implemented four widely used IIS extraction methods:

\begin{itemize}
\item \textbf{Additive Deletion},
\item \textbf{QuickXplain},
\item \textbf{Dual Ray Analysis + Deletion} (applied after LP relaxation of the pseudo-Boolean model),
\item \textbf{Elastic Filter + QuickXplain}.
\end{itemize}

For consistency, we used the following solver backends for each method:
\begin{itemize}
    \item OR-Tools CP-SAT backend was used for Additive-Deletion, QuickXplain, and our Conflict Graph Analysis algorithm.
    \item OR-Tools LP backend was used for Dual Ray Analysis in the context of LP relaxations, HiGHS solver via scipy.optimize.linprog was used for extracting dual rays.
    \item Gurobi was used as the backend for the Elastic Filter method, since the OR-Tools CP-SAT solver returned feasible results for several infeasible instances under that approach.
\end{itemize}

Table~\ref{tab:iis-comparison} summarizes the IIS extraction times (in seconds) for 12 representative testcases using the five methods described above. Each column reports the runtime for a specific method, allowing for direct comparison across approaches.

Table~\ref{tab:cdcl-stats} summarizes detailed statistics collected from our proposed Conflict Graph Analysis algorithm across 15 selected infeasible scheduling instances. These instances were chosen to represent a wide range of problem sizes and structures. Each row corresponds to a distinct test case and reports various internal metrics observed during the IIS extraction process.

The columns include:

\begin{itemize}
    \item \textbf{\#Cons}: The total number of pseudo-Boolean constraints in the original model.
    \item \textbf{\#Vars}: The number of Boolean variables used.
    \item \textbf{AvgLit}: The average number of literals (terms) per constraint.
    \item \textbf{RedCons}: The number of constraints in the final conflict set extracted before minimization.
    \item \textbf{Conflicts}: The number of conflicts encountered during the search.
    \item \textbf{Decisions}: The total number of variable assignments made as decisions.
    \item \textbf{Backtracks}: The number of times the algorithm backtracked to earlier decision levels.
    \item \textbf{Learned}: The number of learned constraints generated via MIR-based conflict resolution.
    \item \textbf{MaxDL}: The highest decision level reached during the search.
    \item \textbf{Time (s)}: The total runtime taken by the algorithm for each instance, measured in seconds.
\end{itemize}

\section{Discussion}

The experimental results show that our conflict graph-based method, when combined with QuickXplain, substantially reduces the time required to compute IISs across a variety of infeasible workforce scheduling problems. Unlike traditional methods such as Additive Deletion or Elastic Filter, which rely heavily on repeated solver calls or model reformulations, our approach uses propagation, implication tracking, and conflict-directed backjumping to isolate conflict sets more efficiently.

While Dual Ray analysis can occasionally achieve faster runtimes, it is limited to models where LP relaxation preserves infeasibility. In contrast, our method remains applicable even when the LP relaxation is feasible but the original pseudo-Boolean system is not, making it more robust for purely Boolean models.

A key advantage of our approach is its ability to quickly extract a small conflict set, thereby reducing the overhead for subsequent minimization using QuickXplain. On average, the final conflict set was \textbf{94.11\%} smaller than the original constraint set, highlighting the effectiveness of early conflict detection and structured search.

The runtime of our algorithm is primarily influenced by the structure of the search tree—specifically the number of conflicts, decision levels, and backtracking steps required to explore infeasible branches. Instances with deeply nested conflicts or many closely interacting policies can trigger numerous decisions and backtracks, increasing runtime. In such cases, the method may perform worse than QuickXplain. However, in most other cases—particularly those involving large constraint sets—our method achieves equal or superior performance. On average, it achieved a \textbf{40\%} reduction in IIS computation time compared to QuickXplain alone.

These results highlight that the effectiveness of conflict graph analysis depends not only on model size but also on the topology of conflicts and propagation paths. Still, the use of efficient unit propagation, conflict-driven learning (via MIR), and backjumping significantly improves scalability and avoids redundant search.

Overall, our method provides a mechanism for explaining infeasibility in complex constraint-based scheduling systems, demonstrating improved performance in many practical scenarios.

\subsection{Regression Analysis of Runtime Behavior}

To better understand the factors influencing the runtime of our Conflict Graph Analysis algorithm, we performed a multiple linear regression using data collected from 50 infeasible test cases. 
The goal of this analysis was to identify which runtime metrics significantly correlate with the total time required to compute the conflict set. We first trained the full regression model using all the above features previously introduced in the Results section, which achieved an $R^2$ value of 0.808, indicating that approximately 81\% of the variance in runtime is explained by these variables. We then performed stepwise refinement:
\begin{enumerate}
    \item We removed less significant variables (\texttt{MaxDL}, \texttt{AvgLit}) based on their high p-values. However, this slightly reduced $R^2$ to 0.805.
    \item One variable (\texttt{RedCons}) had a negative coefficient, though it showed a positive correlation with runtime. This suggests potential multicollinearity.
    \item We computed the Variance Inflation Factor (VIF) for all features and observed extremely high (infinite) VIF scores for \texttt{Conflicts}, \texttt{Decisions}, and \texttt{Backtracks}, indicating strong multicollinearity among them.
    \item Removing these three reduced $R^2$ to 0.689. Reintroducing only \texttt{Backtracks} improved $R^2$ back to 0.808, suggesting it captures the dominant contribution.
\end{enumerate}

Based on refining the initial regression equation, final linear regression model uses the following variables:
\begin{itemize}
    \item \texttt{Total Variables}, \texttt{Constraints}, \texttt{RedCons}, \texttt{Learned}, \texttt{Backtracks}
\end{itemize}

The fitted regression equation is:
\begin{align*}
\text{Time (s)} = -154.94 &+ 0.0045 \cdot (\texttt{Total Variables}) \\
&+ 0.0008 \cdot (\texttt{Total Constraints}) \\
&+ 0.1329 \cdot (\texttt{Backtracks}) \\
&+ 46.7750 \cdot (\texttt{Learned}) \\
&- 1.6518 \cdot (\texttt{RedCons}) \\
&- 1.8644 \cdot (\texttt{MaxDL}) \\
&- 2.3865 \cdot (\texttt{AvgLit})
\end{align*}
For, the above equation, Durbin–Watson statistic was close to 1.9, suggesting minimal autocorrelation in residuals, which also suggests good fit. Overall, this model offers valuable insight into how key runtime factors—particularly \texttt{Backtracks}, \texttt{Learned}, and \texttt{Total Variables}—influence the time taken by the proposed G-CSEA approach.

\begin{table*}[t]
\renewcommand{\arraystretch}{1.4}
\centering
\caption{Comparison of IIS Extraction Methods on Test Instances (Time in Seconds)}
\label{tab:iis-comparison}
\begin{tabular}{lcccccccc}
\toprule
\textbf{Testcase} & \textbf{Total Variables} & \textbf{Total Constraints} & \textbf{1} & \textbf{2} & \textbf{3} & \textbf{4} & \textbf{5} \\
\midrule
1  & 28 & 21 & 0.4172 & 4.6998 & 0.79 & \textbf{0.1972} & 0.3069 \\
2  & 14400 & 3450 & 2193.5713 & 129.426 & 49.18 & \textbf{23.3769} & 30.0481\\
3  & 14400 & 3660 & 1640.3056 & 134.1555 & 56.08 & 232.641 & \textbf{31.4935} \\
4  & 14400 & 3900 & 1528.7273 & 125.0049 & 51.18 & \textbf{28.3186} & 31.8789 \\
5  & 14400 & 4410 & 1754.3504 & 47.328 & 26.92 & \textbf{6.2478} & 20.804 \\
6  & 14400 & 4980 & 1780.712 & 75.751 & 37.60 & \textbf{6.5816} & 22.1212 \\
7  & 480 & 6644 & 36.5488 & 109.26 & -\textsuperscript{*} & \textbf{21.4723} & -\textsuperscript{*} \\
8  & 6720 & 84270 & 294.4949 & 799.2959 & 273.04 & \textbf{6.0994} & 132.3395\\
9  & 14400 & 188460 & 1537.0924 & 1033.0861 & 764.23 & \textbf{13.6644} & 16.2418 \\
10 & 14400 & 199320 & 6753.3965 & 1579.3055 & -\textsuperscript{*} & \textbf{357.0619} & -\textsuperscript{*} \\
11 & 14400 & 199650 & 963.5934 & 1350.3738 & 1267.48 & \textbf{22.3169} & 24.5998 \\
12 & 14400 & 200100 & 1626.5427 & 1338.6396 & 1236.77 & \textbf{20.728} & 21.6033 \\
\bottomrule
\end{tabular}

\vspace{1ex}
\begin{minipage}{0.95\textwidth}
\footnotesize
\textbf{Note:} Columns 1–5 represent different IIS extraction algorithms:\\
\hspace*{1em}%
1: Additive Deletion,\quad
2: QuickXplain,\quad
3: Elastic Filter + QuickXplain,\quad
4: Conflict Graph Analysis + QuickXplain,\quad
5: Dual Ray + Deletion\\
\textsuperscript{*}\,Indicates that it cannot be solved\\
\end{minipage}
\end{table*}

\begin{table*}[t]
\renewcommand{\arraystretch}{1.3}
\caption{Conflict Graph Analysis Statistics Across Selected Test Instances}
\label{tab:cdcl-stats}
\centering
\begin{tabular}{lcccccccccc}
\toprule
\textbf{Testcase} & \textbf{\#Cons} & \textbf{\#Vars} & \textbf{AvgLit} & \textbf{RedCons} & \textbf{Conflicts} & \textbf{Decisions} & \textbf{Backtracks} & \textbf{Learned} & \textbf{MaxDL} & \textbf{Time (s)} \\
\midrule
1 & 21 & 28 & 5.4 & 14 & 54 & 106 & 53 & 2 & 9 & 0.20 \\
2 & 77 & 28 & 2.8 & 28 & 12 & 22 & 11 & 7 & 7 & 0.16 \\
3 & 3660 & 14400 & 13.9 & 14 & 6750 & 13498 & 6749 & 2 & 57 & 232.64 \\
4 & 4410 & 14400 & 11.7 & 3 & 1 & 0 & 0 & 1 & 0 & 6.25 \\
5 & 5790 & 14400 & 9.9 & 18 & 23 & 44 & 22 & 8 & 17 & 9.53 \\
6 & 6644 & 480 & 2.2 & 229 & 15 & 28 & 14 & 15 & 14 & 10.91 \\
7 & 7260 & 14400 & 7.9 & 9 & 16 & 30 & 15 & 4 & 15 & 16.74 \\
8 & 12570 & 28800 & 8.3 & 6 & 1 & 0 & 0 & 1 & 0 & 11.21 \\
9 & 21540 & 28800 & 6.2 & 31 & 3523 & 7044 & 3522 & 6 & 41 & 322.29 \\
10 & 53625 & 72000 & 6.4 & 25 & 3539 & 7076 & 3538 & 7 & 42 & 818.52 \\
11 & 55050 & 72000 & 6.1 & 6 & 1 & 0 & 0 & 1 & 0 & 33.34 \\
12 & 72930 & 6720 & 2.3 & 371 & 70 & 138 & 69 & 20 & 29 & 20.16 \\
13 & 89700 & 6720 & 2.3 & 8 & 1 & 0 & 0 & 1 & 0 & 4.93 \\
14 & 185850 & 14400 & 2.2 & 461 & 255 & 508 & 254 & 29 & 19 & 124.79 \\
15 & 200790 & 14400 & 2.2 & 6 & 1 & 0 & 0 & 1 & 0 & 11.27 \\
\bottomrule
\end{tabular}
\end{table*}


\section{Conclusion}

We presented a conflict graph-based approach for identifying infeasibility in pseudo-Boolean constraint systems, motivated by workforce scheduling models. Our method adapts principles from Conflict-Driven Clause Learning (CDCL) in SAT solvers to extract a conflict set by tracing implication dependencies across decision branches. This conflict set, when minimized using QuickXplain, yields an Irreducible Infeasible Subset (IIS) with significantly fewer solver calls than traditional approaches.

\begin{itemize}
    \item  Improved Efficiency: The conflict graph-based method, when combined with QuickXplain for minimization, significantly reduces the number of solver calls needed to compute Irreducible Infeasible Subsets (IISs) compared to traditional approaches.

   \item Enhanced Performance: Empirical evaluations on various infeasible scheduling instances demonstrate that this approach often outperforms existing techniques such as Additive Deletion, QuickXplain alone, Dual Ray Analysis, and Elastic Filter-based methods. On average, it achieved a 40 percent reduction in IIS computation time compared to QuickXplain alone.

   \item Robustness: Unlike Dual Ray analysis, which is limited to models where LP relaxation maintains infeasibility, the proposed method remains effective even when the LP relaxation is feasible but the original pseudo-Boolean system is not, making it more robust for purely Boolean models.

   \item Compact Conflict Sets: A significant advantage of the approach is its ability to quickly extract a small conflict set, which then reduces the overhead for subsequent minimization. The final conflict set was, on average, 94.11 percent smaller than the original constraint set.

   \item Scalability: The method maintains scalability with problem size and effectively generates compact explanations for infeasibility, even in large, highly constrained systems.
\end{itemize}

\section{Acknowledgement}
The authors acknowledge  Ratnesh Jamidar  and Yoginkumar Patel for their continuous encouragement to drive innovation through research in AI. Special thanks to other Sprinklr AI team members for their  support, and constructive feedback throughout the experiment and evaluation  phase of this work.

\end{document}